\documentclass[twocolumn, 10pt]{article}
\usepackage{graphicx} 
\usepackage[utf8]{inputenc}
\usepackage[T1]{fontenc}
\usepackage{amsmath}  
\usepackage{amssymb}  
\usepackage{amsfonts} 
\usepackage{mathrsfs} 
\usepackage{float} 
\usepackage[a4paper, margin=0.5in]{geometry}
\usepackage{booktabs}
\usepackage{lmodern}
\usepackage{array}
\usepackage{microtype} 
\usepackage[ruled,vlined]{algorithm2e}
\usepackage{subcaption} 
\usepackage{caption}    
\usepackage{xcolor}

\SetKwInOut{Input}{Input}
\SetKwInOut{Output}{Output}

\title{Multi-Class Boundary Extraction from Implicit Representations}
\author{Jash Vira \\
Andrew Myers \\
Simon Ratcliffe\\}
\date{}
\begin{document}
\maketitle

\begin{abstract}
Surface extraction from implicit neural representations modelling a single class surface is a well-known task. However, there exist no surface extraction methods from an implicit representation of multiple classes that guarantee topological correctness and no holes.
In this work, we lay the groundwork by introducing a 2D boundary extraction algorithm for the multi-class case focusing on topological consistency and water-tightness, which also allows for setting minimum detail restraint on the approximation. Finally, we evaluate our algorithm using geological modelling data, showcasing its adaptiveness and ability to honour complex topology.
\end{abstract}

\section{Introduction}
Implicit representations are widely used in computer graphics and vision \cite{sitzmann2020implicit, mescheder2019occupancy} due to their ability to represent data, such as images and complex shapes, in a continuous and differentiable manner. Unlike discrete methods, implicit representations can model fine details and smooth transitions, allowing for evaluation at arbitrary resolutions.

Implicit surface discretisation has been an important area of study with applications ranging from animation to scientific simulation \cite{de2015survey}. However, most existing methods focus on surface extraction from binary-class implicit representations, treating the problem as a simple inside-outside classification \cite{de2015survey, lorensen1998marching}. These approaches are insufficient for geology applications where boundaries shared by multiple classes must be extracted. Such multi-class scenarios require topologically consistent methods to represent complex geological domain models accurately \cite{gonccalves2017machine, hillier2023geoinr}.

In Section~\ref{sec:multiclass}, we introduce a multi-class implicit function in 3D. However, direct surface extraction in 3D poses significant computational challenges and inherent geometric complexity due to multi-class interfaces. Thus, we first establish a rigorous 2D framework as a necessary precursor to handling 3D extraction.
This 2D formulation provides a robust basis for spatially partitioning 3D space into manageable regions. Specifically, the 2D algorithm operates on planar slices of the 3D implicit representation.

This paper's contribution is a novel boundary extraction algorithm for implicit functions modelling multiple classes in 2D. Ensuring topological consistency, a critical challenge in the multi-class setting requires enforcing this property from the earliest stages of the process.
Consequently, we developed a robust one-dimensional root-finding algorithm, described in Section \ref{sec:1drootfinder}, capable of capturing transitions between all represented classes.
Finally, the 2D Polygoniser in section~\ref{sec:2dpolygoniser} details the adaptive spatial partitioning of $\mathbb{R}^2$ into rectangles guided by the 1D root-finder. Using topological and geometrical information derived from the roots found on the edges of these rectangles, our approach constructs a polygonal approximation of the implicit function, ensuring topological accuracy and watertightness.

\section{Multi-Class Implicit Functions}
\label{sec:multiclass}

Let $\mathcal{C} = \{ C_1, C_2, \dots, C_k \}$ be the set of $k$ distinct classes or domains\footnote{The words \textit{class} and \textit{domain} will be used interchangeably throughout the paper.}. For each point $\mathbf{x} \in \mathbb{R}^3$, we seek to determine the corresponding class $C_i \in \mathcal{C}$. 

The training data consists of pairs $(\mathbf{x}, C_i)$, where $\mathbf{x} \in \mathbb{R}^3$ is a point in space, and $C_i \in \mathcal{C}$ is the associated class label. A neural network is trained to approximate the mapping from 3D points to class labels by giving a probability distribution $\mathbf{p}(\mathbf{x}) \in \mathbb{R}^k$, where each element $p_i(\mathbf{x})$ represents the probability that the point $\mathbf{x}$ belongs to class $C_i$.

The class label for each point is determined by selecting the smallest index corresponding to the maximum probability:

\[
C(\mathbf{x}) = \min \left\{ i \in \{1, 2, \dots, k\} \mid p_i(\mathbf{x}) = \max_{j \in \{1, \dots, k\}} p_j(\mathbf{x}) \right\}
\]

A key strength of this approach is that it models the entire 3D space as a differentiable continuous field since the neural network outputs a probability distribution for any point $\mathbf{x} \in \mathbb{R}^3$, unlike discrete voxel-based methods that approximate surfaces at predefined grid points.

\section{1D Root Finder}
\label{sec:1drootfinder}

\begin{figure*}[htb]
    \centering
    \includegraphics[width=1.0\textwidth]{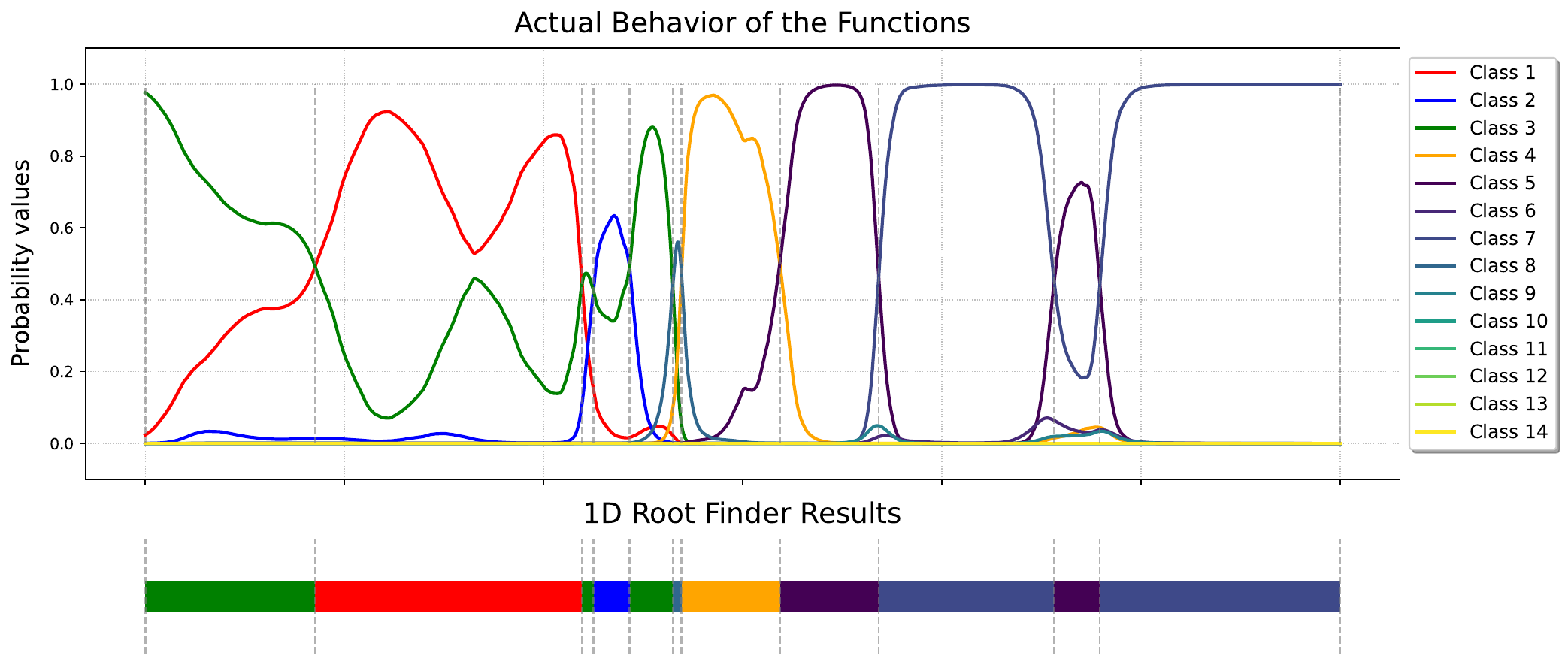}
    \caption{The top part shows the output of a Neural Network for all in $\{ C_1, C_2, \dots, C_k \}$, where $k=14$, densely sampled across $\mathbb{R}^1$. The bottom figure displays the results of the 1D Root-Finder, i.e. the precise intervals where each class with the maximum probability exists.}
    \label{fig:1D_final_result}
\end{figure*}

The 1D Root Finder identifies all domain transitions along an axis-aligned line within an axis-aligned plane. Given two points $x_1, x_2 \in \mathbb{R}$, the objective is to find all transition points $x_r \in [x_1, x_2]$, where a domain switch occurs. 
This method employs an adaptive subdivision strategy guided by a breadth-first search (BFS) approach. We classify an interval $[x_1, x_2]$ based on topological and gradient-based information; each class is described below, followed by the specific details of the bracketing methods and a verification system.

\begin{itemize}
    \item \textbf{Consistent}: If the interval is considered domain-consistent, such an interval goes through a verification mechanism to confirm this finding.
    \item \textbf{Potential Root}: An interval flagged as containing a potential root is first detected by a topological bracketing method, followed by a verification mechanism and then further localised to a single root by a linear-interpolation-based method.
    \item \textbf{Ambiguous}: If the interval's domain cannot be determined, the algorithm subdivides precisely one more level down.
\end{itemize}

By design, the algorithm adaptively subdivides where needed, refining intervals until either the transition points are precisely identified or the interval resolution reaches a predefined threshold.

\subsection{Topological Bracketing}

Traditional root-finding methods are generally effective for binary transitions. However, our problem deals with $k$ distinct classes ($k \geq 2$), where multiple, closely spaced domain transitions can occur along a single axis-aligned line.
This complexity requires a method that accurately identifies all transitions and maintains correct topological ordering.

We introduce a topological bracketing method tailored for multi-class transitions, capable of capturing complex interactions while adapting to the geometry of the transitions.

Let $[x_1, x_2] \subset \mathbb{R}$ be an interval, and consider a neural network function $f: \mathbb{R}^3 \to [0,1]^k$, where for each point $\mathbf{x} \in \mathbb{R}^3$, $f(\mathbf{x}) = [f_1(\mathbf{x}), f_2(\mathbf{x}), \dots, f_k(\mathbf{x})]$ represents the probabilities associated with classes $C_1, C_2, \dots, C_k$, satisfying $\sum_{i=1}^k f_i(\mathbf{x}) = 1$.

At each endpoint $x \in \{x_1, x_2\}$, we rank the classes based on their probabilities $f_i(x)$ in decreasing order. In case of ties (equal probabilities), we prioritise the class with the smallest index. Let $i_1$, $i_2$ denote the indices of the top two classes at $x_1$, and $j_1$, $j_2$ denote the indices of the top two classes at $x_2$.
The classification of the interval $[x_1, x_2]$ is then defined as:
\[
\mathrm{classify}(x_1, x_2) = 
\begin{cases}
    \text{consistent}, & \text{if } i_1 = j_1, \\[0.2cm]
    \text{potential root}, & \text{if } i_1 = j_2 \text{ and } j_1 = i_2, \\[0.2cm]
    \text{ambiguous}, & \text{otherwise}.
\end{cases}
\]

\begin{figure}[!htbp]
    \centering
    \includegraphics[width=0.5\textwidth]{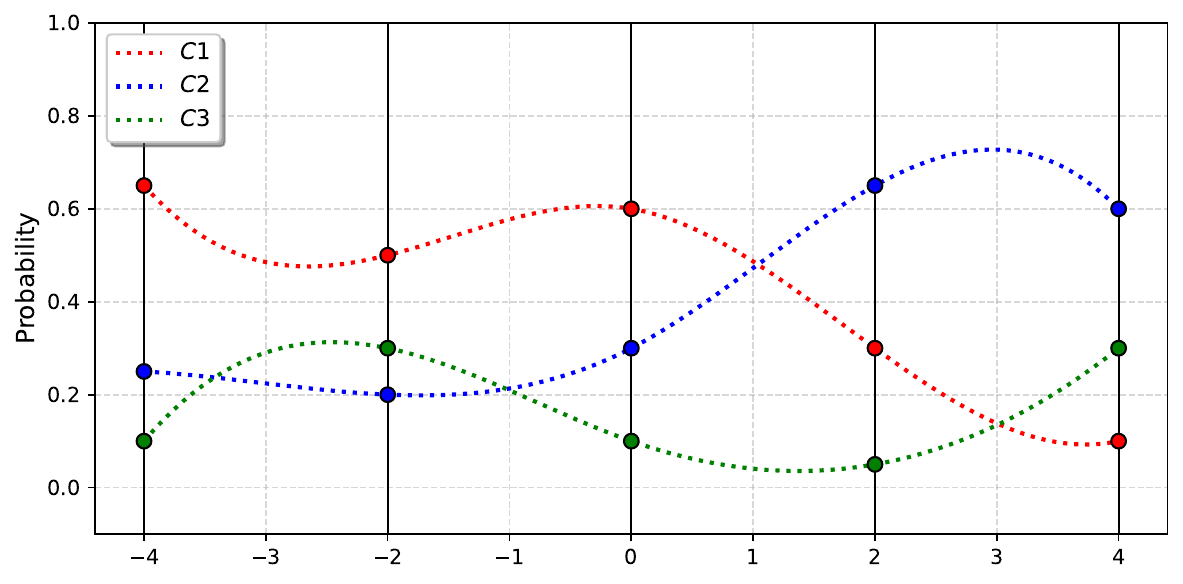}
    \caption{The graph depicts the behaviour of classes $C_1$, $C_2$, and $C_3$ on a 1D axis in the interval $[-4, 4]$. The domain transition occurs in the interval $[0, 2]$, which the 1D algorithm eventually localises using the topological bracketing method.}
    \label{fig:topological_bracketing}
\end{figure}

In Figure~\ref{fig:topological_bracketing}, we can observe how the topological bracketing method will help localise the domain transition. Starting with the interval $[-4, 4]$, which is an \texttt{ambiguous} case and is further bisected. The first child interval $[-4, 0]$ appears \texttt{consistent}, whereas the $[0, 4]$ is \texttt{ambiguous} again. When $[0, 4]$ is bisected further we have $[2, 4]$ which appears \texttt{consistent} and $[0, 2]$ which appears to have a \texttt{potential root} signalling a domain transition.

\subsection{Gradient-Based Localisation}

We introduce a gradient-based linear projection technique to further localise roots identified as \emph{potential} by the topological bracketing method. This method refines the root's position before applying precise root-finding algorithms such as the Newton--Raphson or bisection methods. It effectively handles challenges like multiple roots within an interval and mitigates issues arising from high-frequency variations in the neural network outputs.

Given an interval $[x_1, x_2]$ containing a potential root between two contending classes $C_a$ and $C_b$, we define the function:
\[
g(x) = f_{C_a}(x) - f_{C_b}(x),
\]
where $f_{C_a}(x)$ and $f_{C_b}(x)$ are the probabilities assigned by the neural network $f$ to classes $C_a$ and $C_b$, respectively.

We approximate $g(x)$ near $x_1$ and $x_2$ using linear expansions:
\[
\begin{aligned}
g(x) &\approx g(x_1) + g'(x_1)(x - x_1), \\
g(x) &\approx g(x_2) + g'(x_2)(x - x_2),
\end{aligned}
\]
where $g'(x)$ denotes the derivative of $g$ with respect to $x$.

Solving $g(x) = 0$ for each approximation yields projected root estimates:
\[
\begin{aligned}
x_1^* &= x_1 - \frac{g(x_1)}{g'(x_1)}, \\
x_2^* &= x_2 - \frac{g(x_2)}{g'(x_2)}.
\end{aligned}
\]
These projections indicate where the tangents at $x_1$ and $x_2$ intersect the $x$-axis, providing refined estimates of the root location.

We define $\delta = x_2 - x_1$ as the width of the interval. To assess the likelihood of a root within $[x_1, x_2]$, we consider the following criteria:

\begin{itemize}
    \item \textbf{Definite Root Within Interval}: If both projected points satisfy $x_1 \leq x_i^* \leq x_2$ for $i = 1, 2$, and there is a sign change over the interval, i.e.,
    \[
    g(x_1) \cdot g(x_2) < 0,
    \]
    a root is highly likely within $[x_1, x_2]$. We proceed with the Newton--Raphson method, starting from $x_1$, due to the high confidence in the root's presence.

    \item \textbf{Possible Root Near Interval}: If at least one of the projected points satisfies $x_i^* \in [x_1 - \delta, x_2 + \delta]$ for $i = 1, 2$, or if there is a sign change ($g(x_1) \cdot g(x_2) < 0$), a root may be present near the interval boundaries.

    \item \textbf{Unlikely Root}: If neither projected point lies within $[x_1 - \delta, x_2 + \delta]$ and there is no sign change (i.e., $g(x_1) \cdot g(x_2) > 0$), it is unlikely that a root exists within $[x_1, x_2]$. 
\end{itemize}

\begin{figure}[!htbp]
    \centering
    \includegraphics[width=0.5\textwidth]{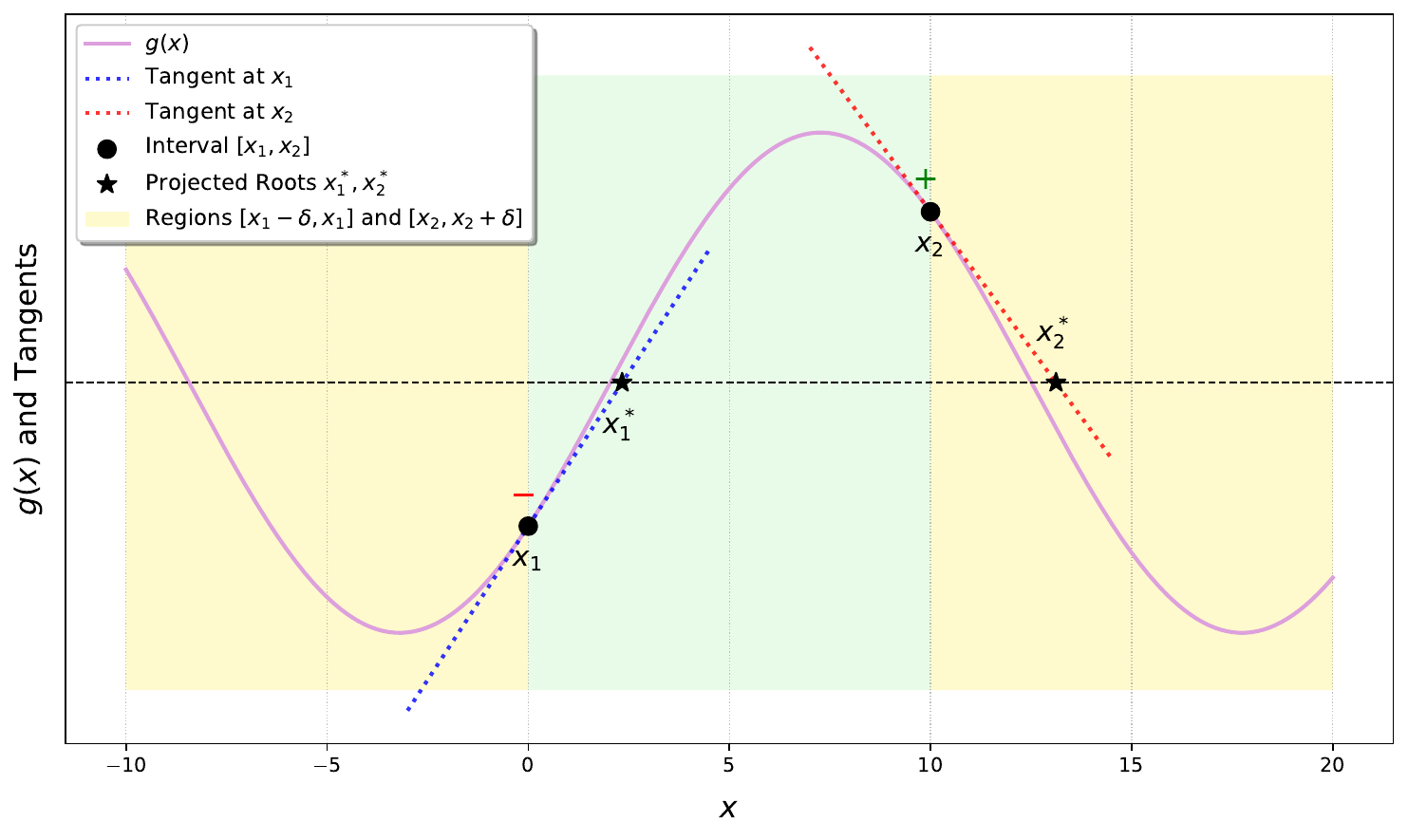}
    \caption{
    Illustration of the gradient-based localisation approach for root finding. The figure demonstrates the scenario where $g(x)$ exhibits multiple domain transitions, with tangents plotted at $x_1$ and $x_2$. The regions $[x_1 - \delta, x_1]$ and $[x_2, x_2 + \delta]$ are shown in yellow, while the interval $[x_1, x_2]$ is in green, indicating the interval under consideration for root localisation. The plot captures a \textit{potential root} classification since the signs of $g(x)$ change at $x_1$ and $x_2$, and not both projected intersections $x_1^*$ and $x_2^*$ lie in the green area.}
    \label{fig:gradient_localisation}
\end{figure}

\subsection{Verification of interval}

The verification technique confirms deterministic classifications before advancing to subsequent stages of the algorithm. The idea is to ensure the reliability of classifications, which is crucial for accurate surface extraction from implicit functions.

Verification is based on recursive subdivision, where the parameter $V_n \in \mathbb{N}$ represents the number of subdivision levels used for verification. In the case of $\mathbb{R}^1$, we take up the examples of the \emph{consistent} and \emph{potential root} classifications.

For a node classified as a \emph{potential root}, verification proceeds as follows:

We recursively subdivide the original node $V_n$ levels down. This process creates $2^{(V_n+1)} - 2$ new nodes. The verification of this subtree involves the following checks:

\begin{itemize}
    \item A single leaf node maintains the \emph{potential root} classification.
    \item All other leaf nodes are classified as \emph{consistent}, ensuring no contradictions and isolating the \emph{potential root} nodes.
\end{itemize}

For a node classified as \emph{consistent}, the verification involves recursively subdividing the node for $V_n$ levels and ensuring that all resulting subnodes maintain the \emph{consistent} classification. This ensures homogeneity across subdivisions.

This general approach can be applied to $\mathbb{R}^1$, $\mathbb{R}^2$, and $\mathbb{R}^3$ dimensions, making it adaptable for various contexts throughout this work.

\subsection{Algorithm Overview}

Algorithm~\ref{alg:1D_root_finder} in Appendix A presents a one-dimensional root-finding algorithm that integrates the topological bracketing method, gradient-based localisation, and verification techniques. Operating on an axis-aligned interval, the algorithm employs a breadth-first search strategy to locate roots efficiently. It prioritises intervals likely to yield definitive results through a verification queue, enhancing the reliability of classifications before applying computationally intensive methods like gradient-based localisation. The Interval Limit ($\epsilon$) is typically set to approximate the smallest representable difference in the chosen floating-point format. 

\section{2D Polygoniser}
\label{sec:2dpolygoniser}

The 2D Polygoniser reconstructs the geometry within an axis-aligned rectangle in $\mathbb{R}^2$. This is the fundamental unit of analysis in 2D. The algorithm operates by examining these rectangles and applying an adaptive subdivision strategy to resolve the geometry within them. The process leverages topological information, mainly focusing on domain transitions along the edges of the rectangle.

\subsection{2D Classes and Subdivision}
\label{sec:2dclasses}
The 1D root-finder is applied along each rectangle edge to identify transition points where the domain classification changes. If any edge contains more than one transition point, the rectangle is immediately classified as \textbf{ambiguous} and marked for subdivision.

For rectangles where each edge has at most one transition point, the classification is determined based on the domain labels at the vertices and the positions of the transition points along the edges. The possible classifications include:

\begin{itemize}
    \item \textbf{Polygonisable}: The rectangle's topology is clear and can be directly polygonised without further subdivision.
    \item \textbf{Ambiguous}: The topology within the rectangle is still uncertain, requiring subdivision for resolution.
    \item \textbf{Three Domains Meeting (TDM)}: The topology indicates three distinct domains meet at a point within the rectangle, necessitating special handling.
\end{itemize}

\begin{figure}[!tb]
    \centering
    \captionsetup[subfigure]{skip=-2pt}
    \renewcommand{\arraystretch}{0.8} 

    \begin{tabular}{|>{\centering\arraybackslash}m{2.25cm}>{\centering\arraybackslash}m{2cm}>{\centering\arraybackslash}m{2.25cm}|}
        \hline
        \multicolumn{3}{|c|}{\textbf{\normalsize Polygonisable}} \\ \hline
        
        \subcaptionbox{}{\includegraphics[width=2.0cm]{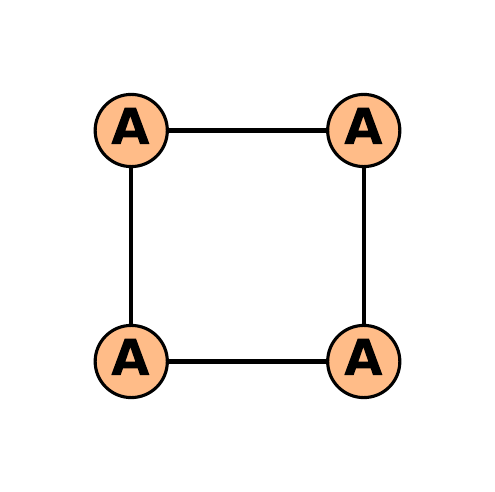}} &
        \subcaptionbox{}{\includegraphics[width=2.0cm]{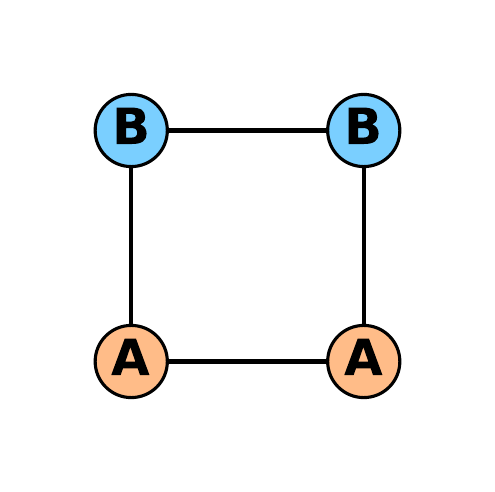}} &
        \subcaptionbox{}{\includegraphics[width=2.0cm]{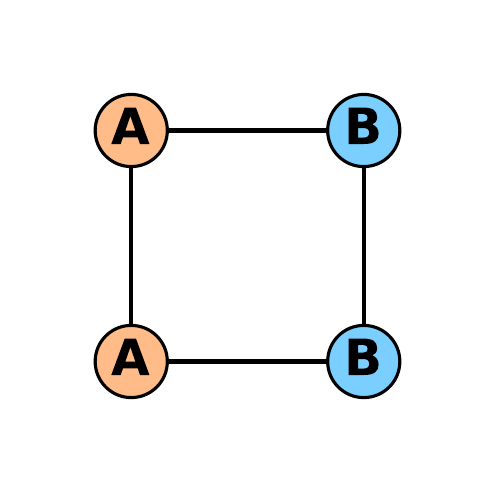}} \\[2pt]
        
        \subcaptionbox{}{\includegraphics[width=2.0cm]{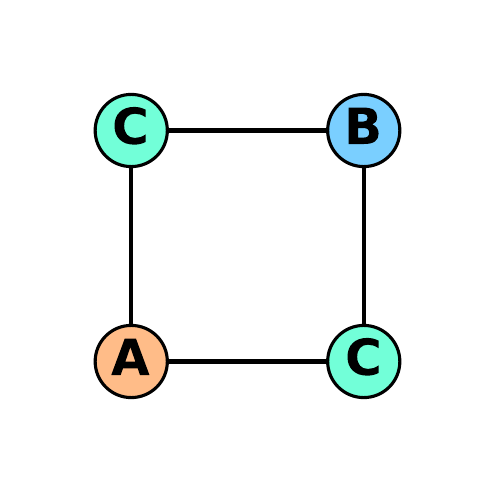}} &
        \subcaptionbox{}{\includegraphics[width=2.0cm]{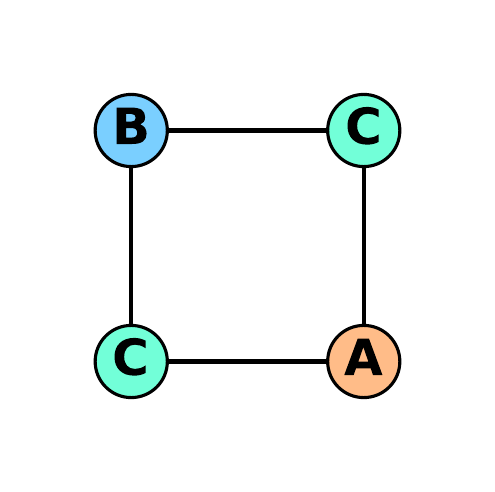}} &
        \subcaptionbox{}{\includegraphics[width=2.0cm]{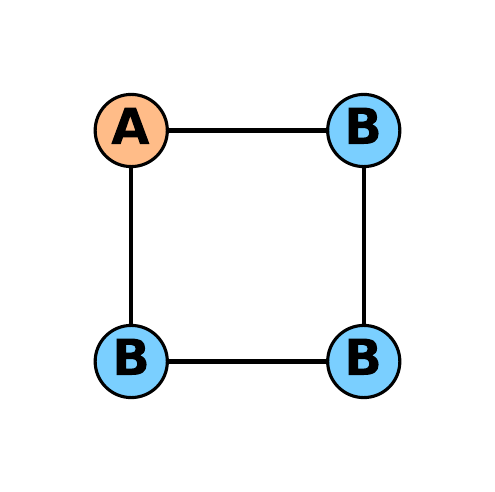}} \\[2pt]
        
        \subcaptionbox{}{\includegraphics[width=2.0cm]{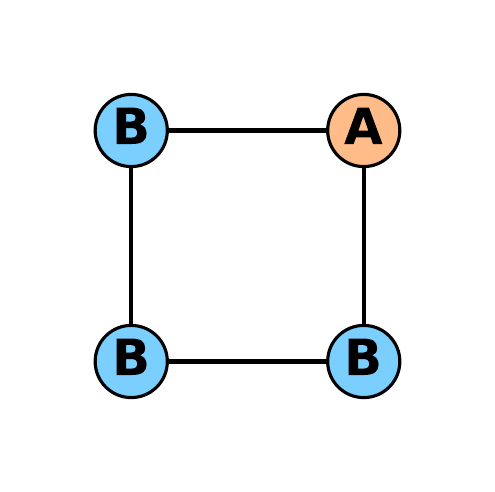}} &
        \subcaptionbox{}{\includegraphics[width=2.0cm]{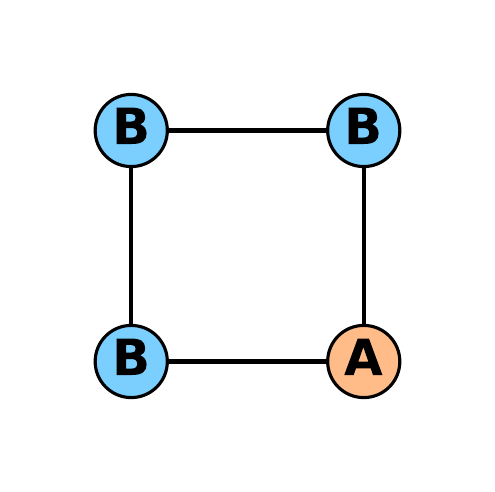}} &
        \subcaptionbox{}{\includegraphics[width=2.0cm]{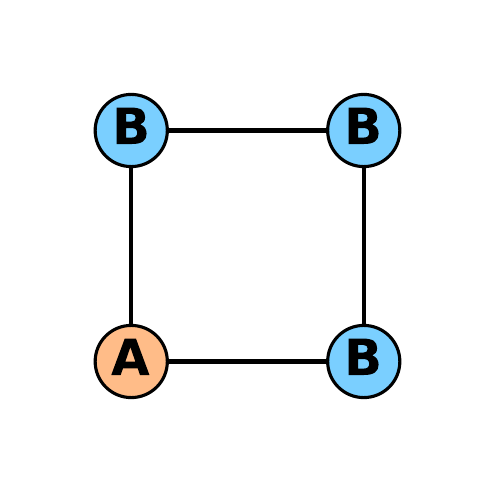}} \\ \hline
    \end{tabular}

    \vspace{-0.05cm}

    \begin{tabular}{|>{\centering\arraybackslash}m{2.25cm}>{\centering\arraybackslash}m{2cm}|>{\centering\arraybackslash}m{2.25cm}|}
        \hline
        \multicolumn{2}{|c|}{\textbf{\normalsize Three Domains Meeting}} & \textbf{\normalsize Ambiguous} \\ \hline
        
        \subcaptionbox{}{\includegraphics[width=2.0cm]{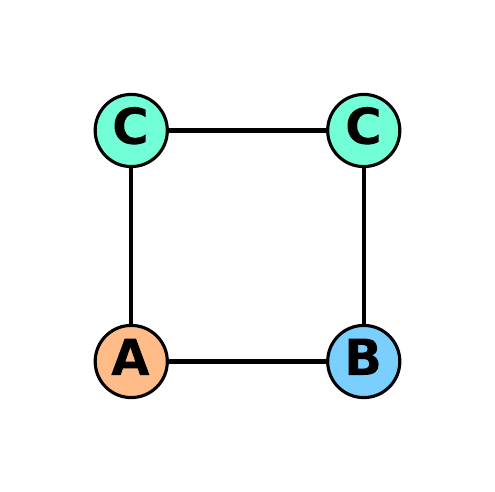}} &
        \subcaptionbox{}{\includegraphics[width=2.0cm]{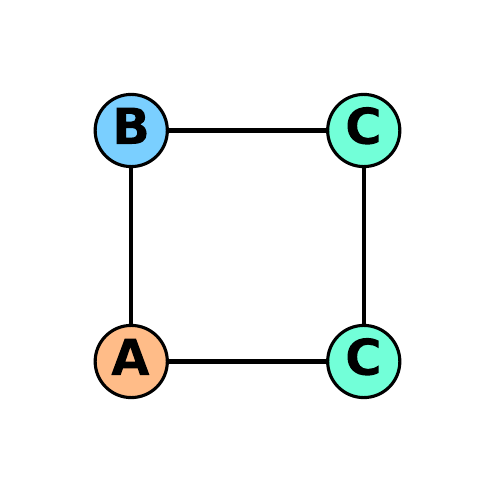}} &
        \subcaptionbox{}{\includegraphics[width=2.0cm]{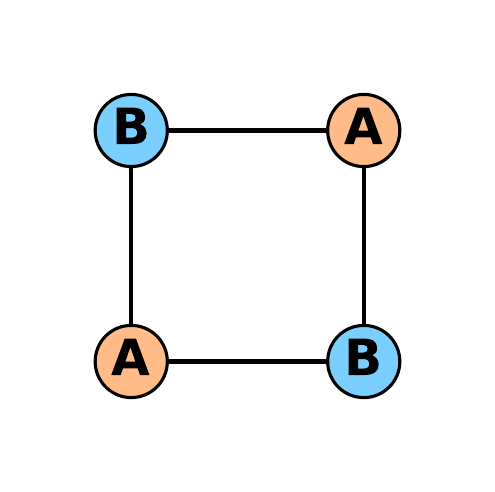}} \\[2pt]
        
        \subcaptionbox{}{\includegraphics[width=2.0cm]{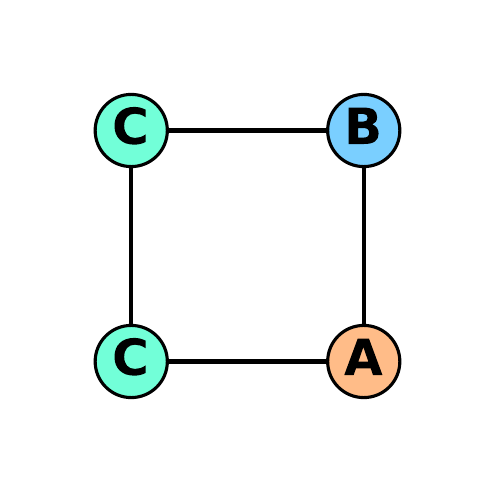}} &
        \subcaptionbox{}{\includegraphics[width=2.0cm]{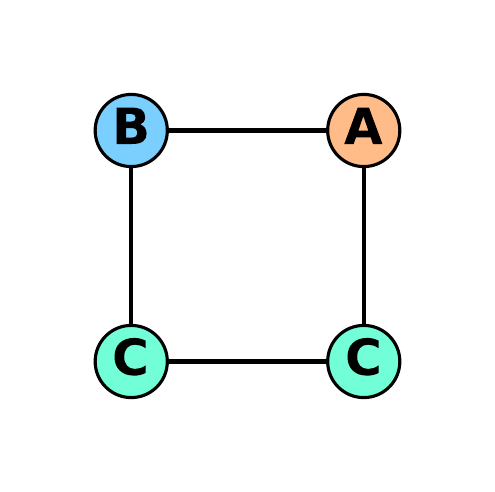}} &
        \subcaptionbox{}{\includegraphics[width=2.0cm]{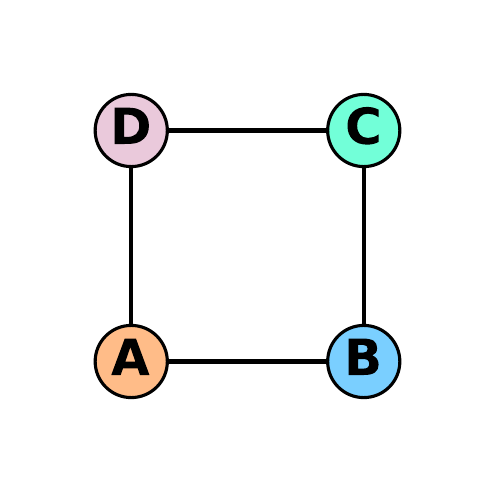}} \\ \hline
    \end{tabular}

    \caption{This figure showcases all possible permutations of the Polygonisable, Three Domains Meeting and Ambiguous cases of a 2D rectangle.}
    \label{fig:quadrilateral_classes}
\end{figure}

Figure~\ref{fig:quadrilateral_classes} details specific configurations and their corresponding classifications.

Unlike simple bisection in 1D, 2D subdivision alternates between the $x$ and $y$ directions. Let $d \in \{x, y\}$ denote the current subdivision direction. The set of transition points is defined as:
\[
\begin{aligned}
R_d = \{r_i \in \mathbb{R} \mid r_i = \pi_d(p), \, \\
p \text{ is a root on the edges parallel to } d\}
\end{aligned}
\]
where $\pi_d$ denotes the projection onto the $d$-axis. The points $r_i$ are sorted such that $r_1 < r_2 < \cdots < r_n$, with $n = |R_d|$.

The midpoints of consecutive transition points are given by:
\[
M = \left\{ m_i = \frac{r_i + r_{i+1}}{2} \mid i = 1, \ldots, n-1 \right\}.
\]

The subdivision point $m^*$ is selected as:
\[
m^* = 
\begin{cases} 
    m_{\lfloor |M|/2 \rfloor + 1}, & \text{if } |M| \text{ is odd}, \\
    \arg\max_{i \in \{\lfloor |M|/2 \rfloor, \lfloor |M|/2 \rfloor + 1\}} \Delta(m_i), & \text{if } |M| \text{ is even},
\end{cases}
\]
where $\Delta(m_i)$ is the interval size associated with $m_i$.

Based on the limited information available from its boundaries, the selected subdivision point $ m*$ represents an area of high topological change in the geometry within the rectangle.

\subsection{Triple Junction Estimation}
\label{sec:2dtriplejunction}

Consider three scalar fields $f_1, f_2, f_3 : \mathbb{R}^2 \to \mathbb{R}$. A \emph{triple junction} is a point $(x^*, y^*)$ at which
\[
f_1(x^*, y^*) = f_2(x^*, y^*) = f_3(x^*, y^*).
\]

When a rectangle is classified as containing such a triple junction, we seek $(x^*, y^*)$ to a prescribed tolerance $\varepsilon > 0$. The estimation proceeds in two stages: an iterative tangent-plane method followed by a fallback optimisation-based approach if the former fails.

\paragraph{Tangent-Plane Method}
At an iterate $(x, y)$, let $f_i(x, y) = z_i$ and $\nabla f_i(x, y) = (a_i, b_i)^\top$. Approximating each $f_i$ linearly about $(x,y)$ gives:
\[
f_i(X,Y) \approx z_i + a_i(X - x) + b_i(Y - y).
\]
We seek $(X,Y)$ that makes $f_1(X,Y)$ equal to $f_2(X,Y)$ and $f_3(X,Y)$ simultaneously. However, instead of directly enforcing these equalities pairwise, we introduce an auxiliary variable $Z$ and consider the linear systems of the form:
\[
a_i X + b_i Y - Z = a_i x + b_i y - z_i, \quad i = 1,2,3.
\]

If a solution $(X,Y,Z)$ exists, it represents a configuration where three approximate tangent surfaces (one for each $f_i$) intersect consistently. In practice, we solve these three linear equations in a least-squares sense, obtaining $(X,Y,Z)$ that minimizes any residual inconsistency. We then update the iterate:
\[
(x,y) \leftarrow (X,Y),
\]
this repeats until $\sqrt{(X - x)^2 + (Y - y)^2} < \varepsilon$ or a maximum iteration count is reached. If this yields $(x^*,y^*)$ inside the rectangle with the desired precision, we accept $(x^*,y^*)$.

\paragraph{Optimisation-Based Fallback}
If the tangent-plane method does not converge within the rectangle, we minimise the objective function:
\[
G(x, y) = \sum_{1 \leq i < j \leq 3} \big(f_i(x, y) - f_j(x, y)\big)^2.
\]
The optimisation problem is:
\[
(x^*, y^*) = \arg\min_{(x, y) \in \text{bounds}} G(x, y).
\]
Using a derivative-free optimisation algorithm (e.g., Nelder–Mead), we accept $(x^*, y^*)$ if:
\[
G(x^*, y^*) < \varepsilon \quad \text{and} \quad (x^*, y^*) \in \text{bounds}.
\]
Where the bounds are the extent of the rectangle.

This two-stage approach ensures robustness by initially leveraging gradient information for efficient convergence and subsequently employing a general optimisation strategy to guarantee solution reliability.

\subsection{2D Verification}
\label{sec:2dverification}

Recall from Subsection~2.3 that in the 1D root-finder, descending $V_n$ levels generates $2^{(V_n+1)} - 2$ new nodes. The same principle applies in two dimensions, with the distinction that subdivision occurs at a non-deterministic position, as governed by the method outlined in Subsection~3.1.

To verify the \textit{Polygonisable} class in two dimensions, all nodes in its verification subtree must maintain the same classification. For the \textit{TMD} class, exactly one leaf node must exhibit the triple-domain meeting condition, while all remaining leaf nodes are classified as \textit{Polygonisable}.

\subsection{Geometric Criteria}
\label{sec:2dgeometriccriterion}

The geometric criteria refine the polygonisation process by selectively subdividing edges that fail to capture local geometric complexity. Verified edges are part of rectangles classified as either \emph{Polygonisable} or \emph{TDM}, where each edge approximates the contour $f=0$, defined as the scalar field difference $f=f_{c_1}-f_{c_2}$ between two classes meeting along the edge.

Consider a verified edge with endpoints $(x_1, y_1)$ and $(x_2, y_2)$. At each endpoint, the gradient of $f$ is evaluated as $\nabla f(x_i, y_i) = (a_i, b_i)$ for $i=1,2$. The tangent lines at these endpoints are:
\[
a_i (x - x_i) + b_i (y - y_i) = 0, \quad i=1,2.
\]
Solving these equations provides the intersection point $(x_I, y_I)$, which represents the intersection of the tangent approximations to the contour $f=0$ at the edge endpoints.

Let \( L \) denote the length of the edge connecting \((x_1, y_1)\) and \((x_2, y_2)\), and let \(\delta > 0\) be a user-defined geometric threshold. Let \( d \) be the perpendicular distance from \((x_I, y_I)\) to the line segment connecting \((x_1, y_1)\) and \((x_2, y_2)\). If both \( d > \delta \) and \( L > \delta \), the rectangle is subdivided to enhance the fidelity of the approximation.

In the case of subdivision, the rectangle is divided precisely at the midpoint of the edge being evaluated, either along the $x$-axis or $y$-axis, depending on the edge's orientation. This adaptive criterion ensures that regions with high geometric complexity are refined while simpler areas remain coarsely represented.

\begin{figure}[!htbp]
    \centering
    \includegraphics[width=0.5\textwidth]{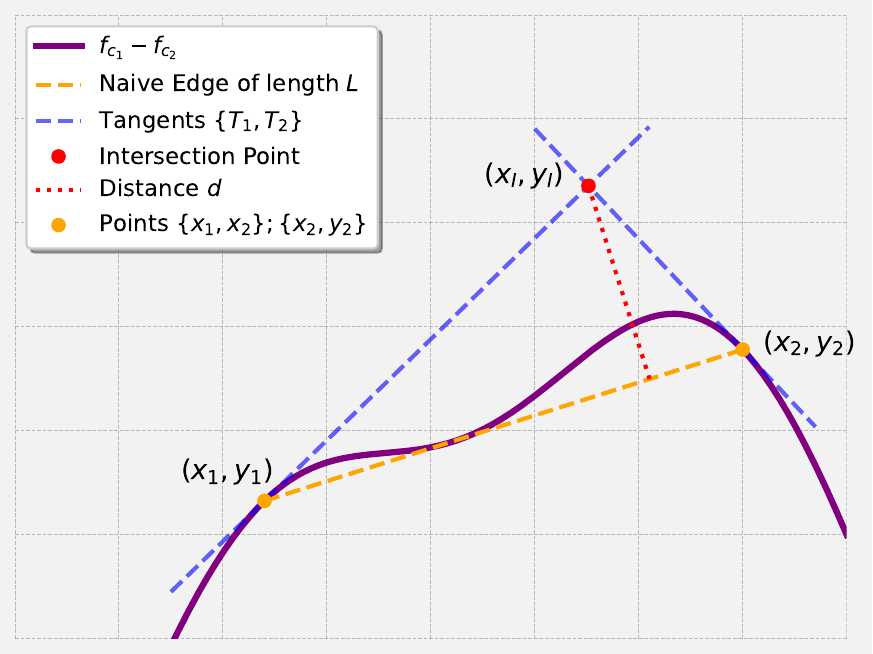}
    \caption{Illustration of method imposing geometric criteria on a naive edge represented in yellow, which is trying to approximate the boundary in purple.}
    \label{fig:geometric_criterion_plot}
\end{figure}

\subsection{2D Algorithm Overview}

In Algorithm~\ref{alg:2d_polygoniser}, we outline the adaptive subdivision mechanism employed to partition a rectangle in $\mathbb{R}^2$ into regions of recognisable topological cases. This is followed by systematically refining the geometry to a desired level of detail, finally producing a network of edges representing the boundaries between classes.

\begin{enumerate}
    \item \textbf{Classification}:  
    The algorithm begins by initialising two stacks: a \textit{verification stack} $\mathcal{V}$ for rectangles classified as either \texttt{Polygonisable} or \texttt{TDM}, and a \textit{normal stack} $\mathcal{S}$ for ambiguous rectangles. These classifications are detailed in Subsection~\ref{sec:2dclasses}. An empty edge store $\mathcal{E}$ is prepared to collect the final edge network and its associated class transitions.

    \item \textbf{Processing Stacks}:  
    The verification stack $\mathcal{V}$ is prioritised over the normal stack $\mathcal{S}$:
    \begin{itemize}
        \item \textbf{Verification Stack}:  
        Rectangles in $\mathcal{V}$ undergo verification using \texttt{verify\_rectangle}, as described in Subsection~\ref{sec:2dverification}. For \texttt{TDM} rectangles, the triple junction is estimated using \texttt{find\_triple\_junction} (Subsection~\ref{sec:2dtriplejunction}) before applying the geometric criterion (\texttt{apply\_geometric\_criterion}, Subsection~\ref{sec:2dgeometriccriterion}) and connecting edges with \texttt{connect\_edges}.
        
        \item \textbf{Normal Stack}:  
        Rectangles in $\mathcal{S}$ are subdivided using \texttt{subdivide\_rectangle}, as detailed in Subsection~\ref{sec:2dclasses}. Child rectangles are classified, with \texttt{Polygonisable} and \texttt{TDM} children added to $\mathcal{V}$, and \texttt{ambiguous} children added to $\mathcal{S}$.
    \end{itemize}

\end{enumerate}

\begin{algorithm}[htbp]
\footnotesize
\caption{Two-Dimensional Polygonisation}
\Input{
    Rectangle extents $R_{\text{root}}$, verification depth $V_n$, geometric criteria threshold $\delta$, subdivision limit $\epsilon$
}
\Output{
    Edge network $\mathcal{E}$
}

\SetKwProg{Fn}{def}{:}{}
\SetKwFunction{Classify}{classify\_rectangle}
\SetKwFunction{Subdivide}{subdivide\_rectangle}
\SetKwFunction{Verify}{verify\_rectangle}
\SetKwFunction{Criterion}{apply\_geometric\_criterion}
\SetKwFunction{Connect}{connect\_edges}
\SetKwFunction{TripleJunction}{find\_triple\_junction}
\SetKwFunction{Push}{push}
\SetKwFunction{Pop}{pop}
\SetKwIF{If}{ElseIf}{Else}{if}{:}{elif}{else:}{}%
\SetKwFor{While}{while}{:}{}
\SetKwFor{ForEach}{for}{in}{}
\SetAlgoNoEnd

\Fn{2D Polygoniser($R_{\text{root}}, V_n, \delta, \epsilon$)}{
    \textbf{Initialise:} \\
    Normal stack $\mathcal{S} \leftarrow \emptyset$, Verification stack $\mathcal{V} \leftarrow \emptyset$, Edge store $\mathcal{E} \leftarrow \emptyset$ \\
    \BlankLine

    $C \gets \Classify(R_{\text{root}})$ \\
    \If{$C \in \{\texttt{Polygonisable}, \texttt{TDM}\}$}{
        $\mathcal{V}.\Push(R_{\text{root}})$ \tcp*[r]{Add to verification stack}
    }
    \Else{
        $\mathcal{S}.\Push(R_{\text{root}})$ \tcp*[r]{Add to normal stack}
    }
    \BlankLine

    \While{$\mathcal{V} \neq \emptyset$ \textbf{or} $\mathcal{S} \neq \emptyset$}{
        \uIf{$\mathcal{V} \neq \emptyset$}{
            $R \gets \mathcal{V}.\Pop()$ \\
            \If{$\Subdivide(R) < \epsilon$}{
                \textbf{continue} \tcp*[r]{Subdivision limit reached}
            }
            $V \gets \Verify(R, V_n)$ \\
            \If{$C = \texttt{TDM}$}{
                $\TripleJunction(R)$ \tcp*[r]{Find triple junction}
            }
            $\Criterion(R, \delta)$ \tcp*[r]{Apply geometric criterion}
            $\mathcal{E} \gets \Connect(R)$ \tcp*[r]{Store edges}
        }
        \Else{
            $R \gets \mathcal{S}.\Pop()$ \\
            \If{$\Subdivide(R) < \epsilon$}{
                \textbf{continue} \tcp*[r]{Subdivision limit reached}
            }
            $R_{\text{1}}, R_{\text{2}} \gets \Subdivide(R)$ \\
            \ForEach{$R_{\text{child}} \in \{R_{\text{1}}, R_{\text{2}}\}$}{
                $C_{\text{child}} \gets \Classify(R_{\text{child}})$ \\
                \If{$C_{\text{child}} \in \{\texttt{Polygonisable}, \texttt{TDM}\}$}{
                    $\mathcal{V}.\Push(R_{\text{child}})$
                }
                \Else{
                    $\mathcal{S}.\Push(R_{\text{child}})$
                }
            }
        }
    }
    \Return{$\mathcal{E}$}
\label{alg:2d_polygoniser}}
\end{algorithm}

\section{Results}

\begin{figure*}[t]
    \centering

    Sampled 1000 x 1000 \\[0.1cm] 

    \begin{minipage}[t]{0.31\textwidth}
        \centering
        \includegraphics[width=\linewidth, height=5cm, trim={0cm 0cm 0cm 0cm}, clip]{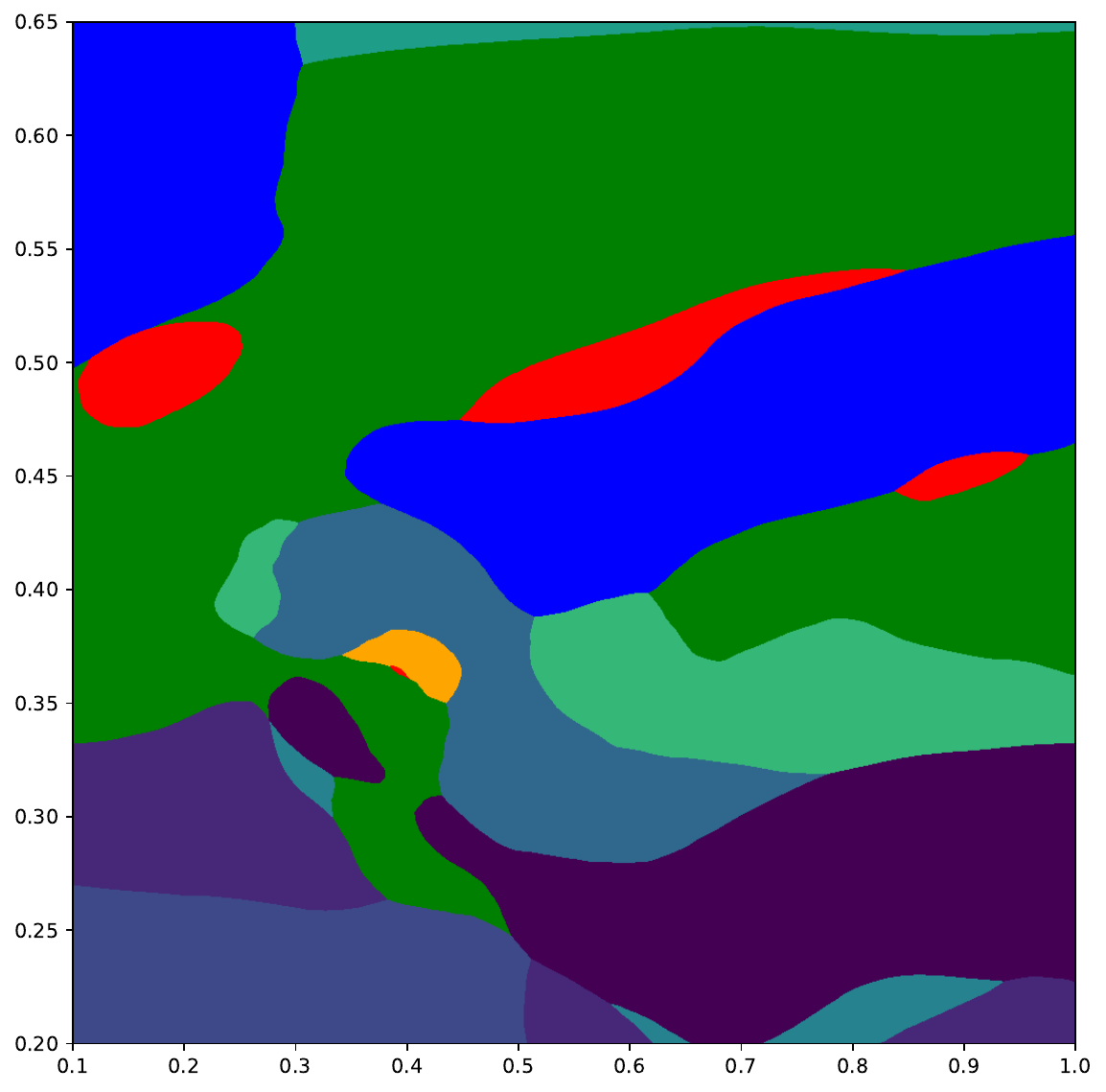}
    \end{minipage}
    \hspace{0.01\textwidth}
    \begin{minipage}[t]{0.31\textwidth}
        \centering
        \includegraphics[width=\linewidth, height=5cm, trim={0cm 0cm 0cm 0cm}, clip]{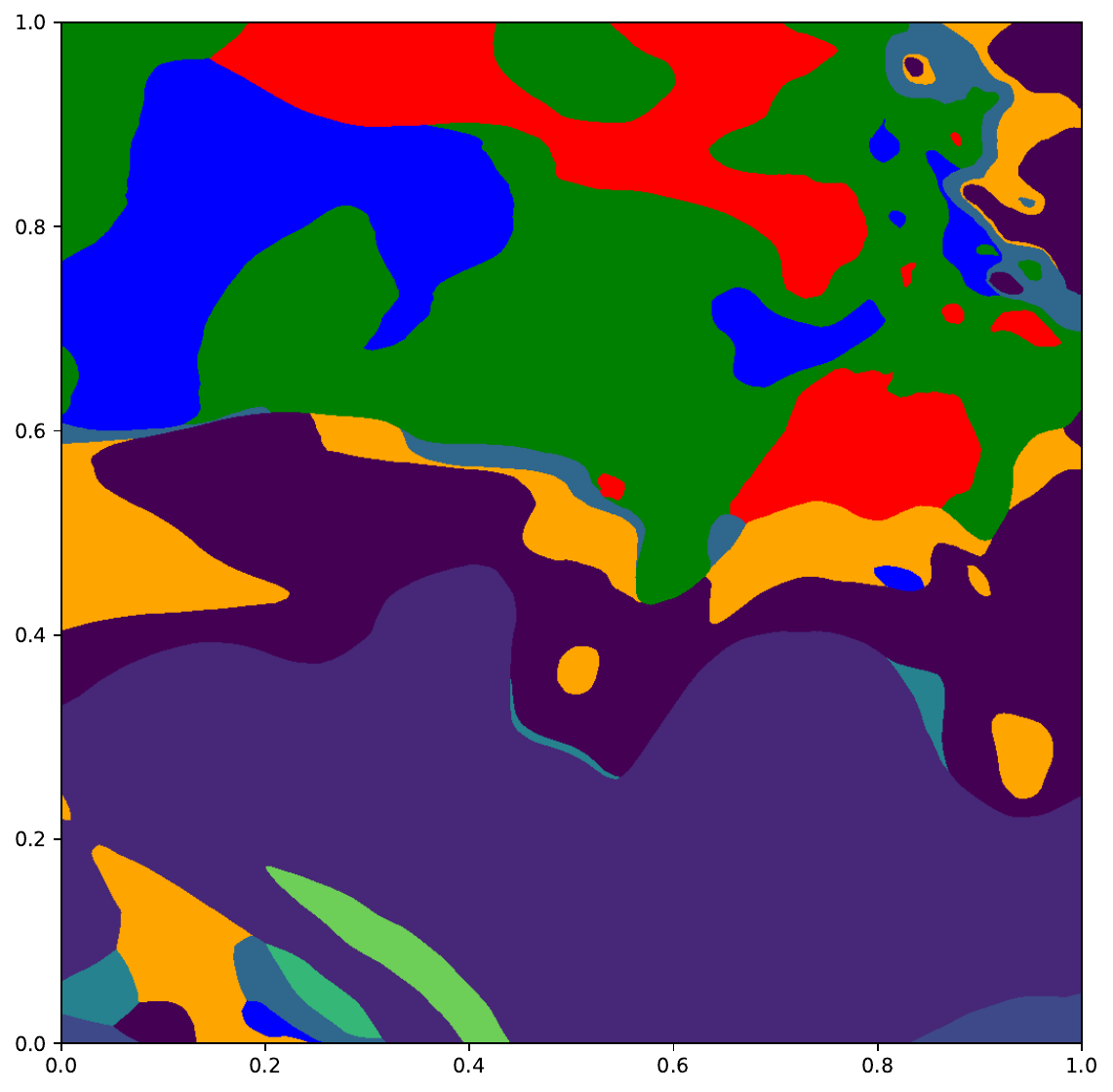}
    \end{minipage}
    \hspace{0.01\textwidth}
    \begin{minipage}[t]{0.31\textwidth}
        \centering
        \includegraphics[width=\linewidth, height=5cm, trim={0cm 0cm 0cm 0cm}, clip]{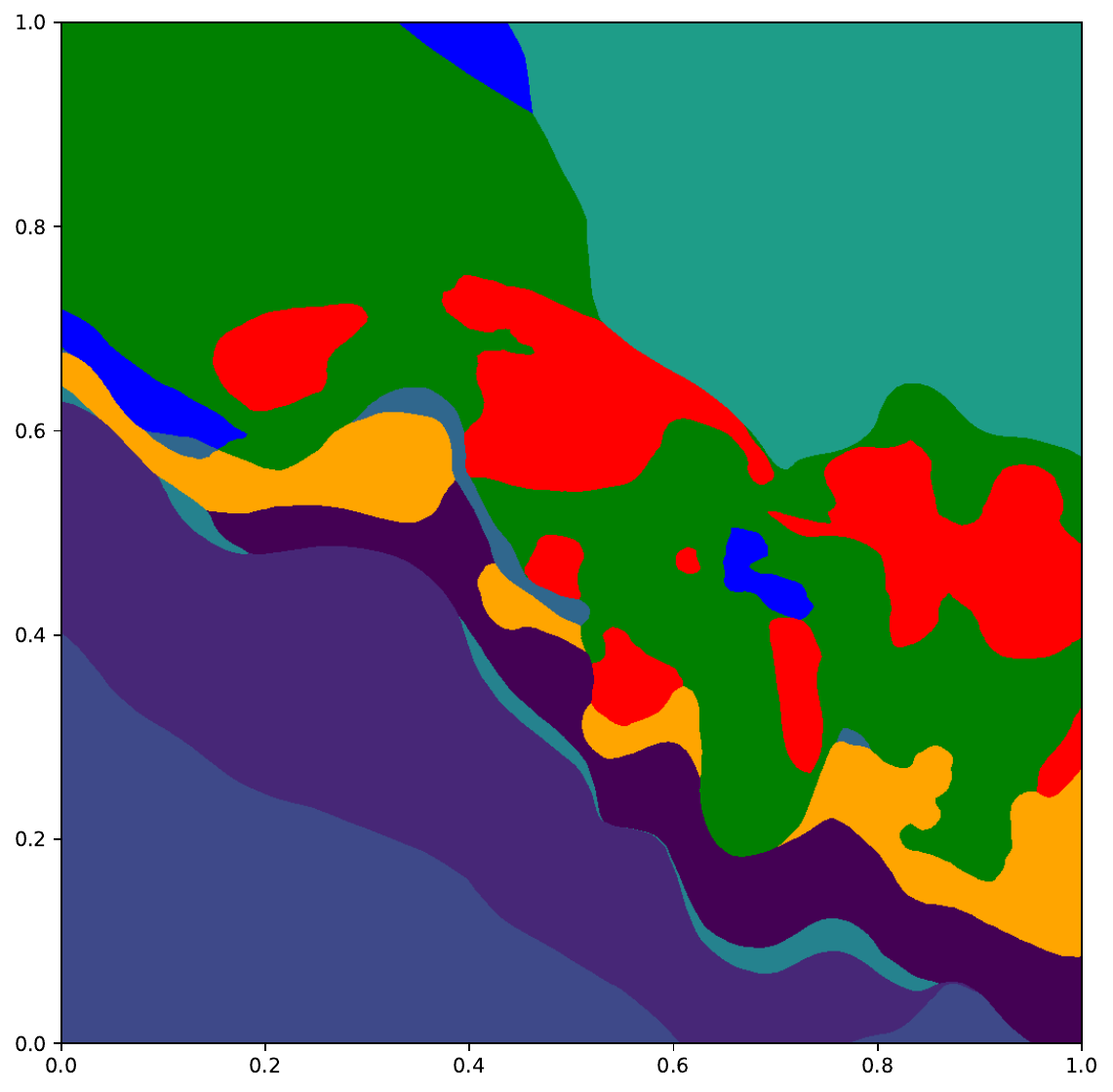}
    \end{minipage}

    \vspace{0.1cm} 

    Extracted boundaries by our algorithm \\[0.1cm] 

    \begin{minipage}[t]{0.31\textwidth}
        \centering
        \includegraphics[width=\linewidth, height=5cm, trim={0cm 0cm 0cm 0cm}, clip]{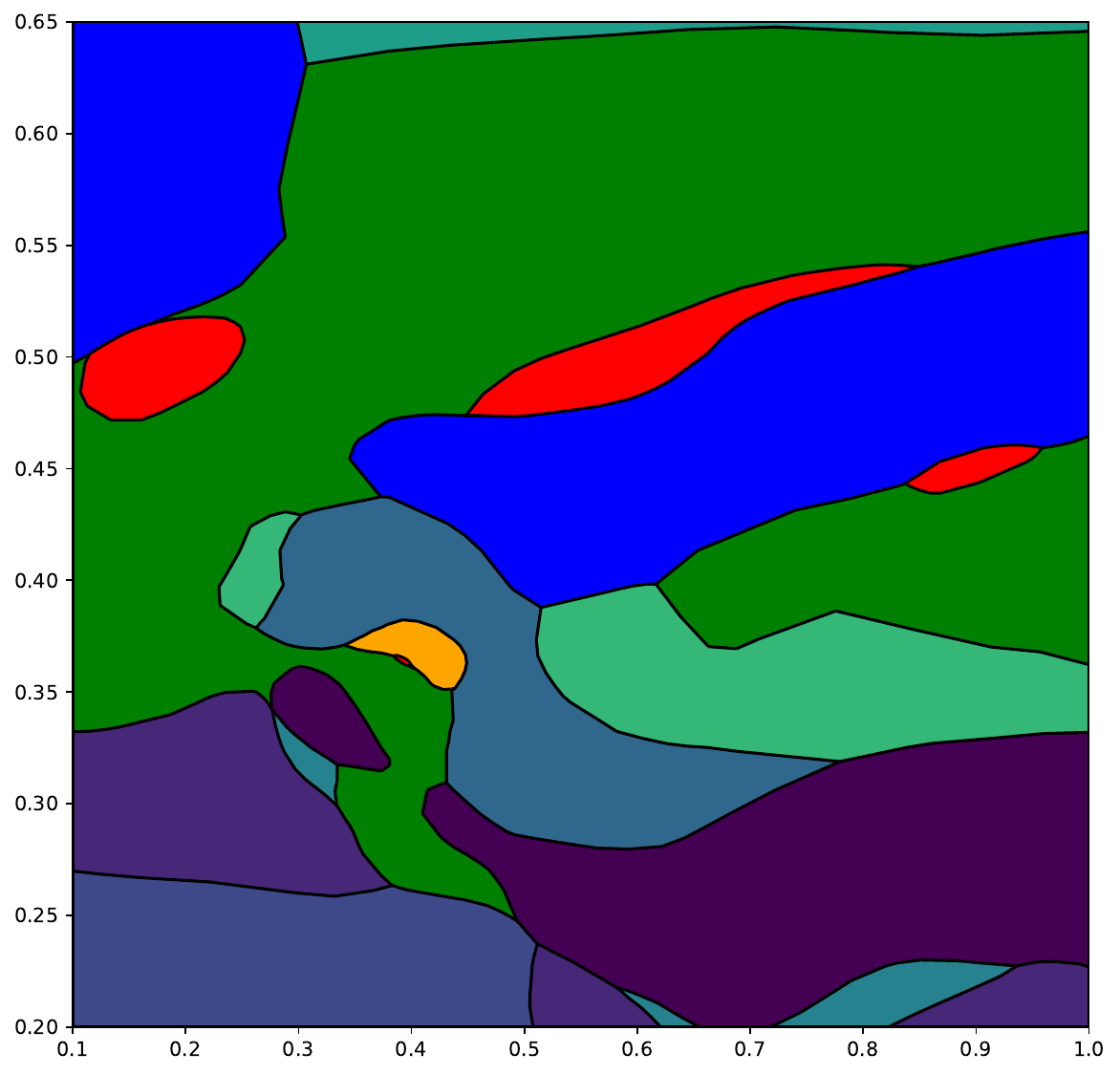}
    \end{minipage}
    \hspace{0.01\textwidth}
    \begin{minipage}[t]{0.31\textwidth}
        \centering
        \includegraphics[width=\linewidth, height=5cm, trim={0cm 0cm 0cm 0cm}, clip]{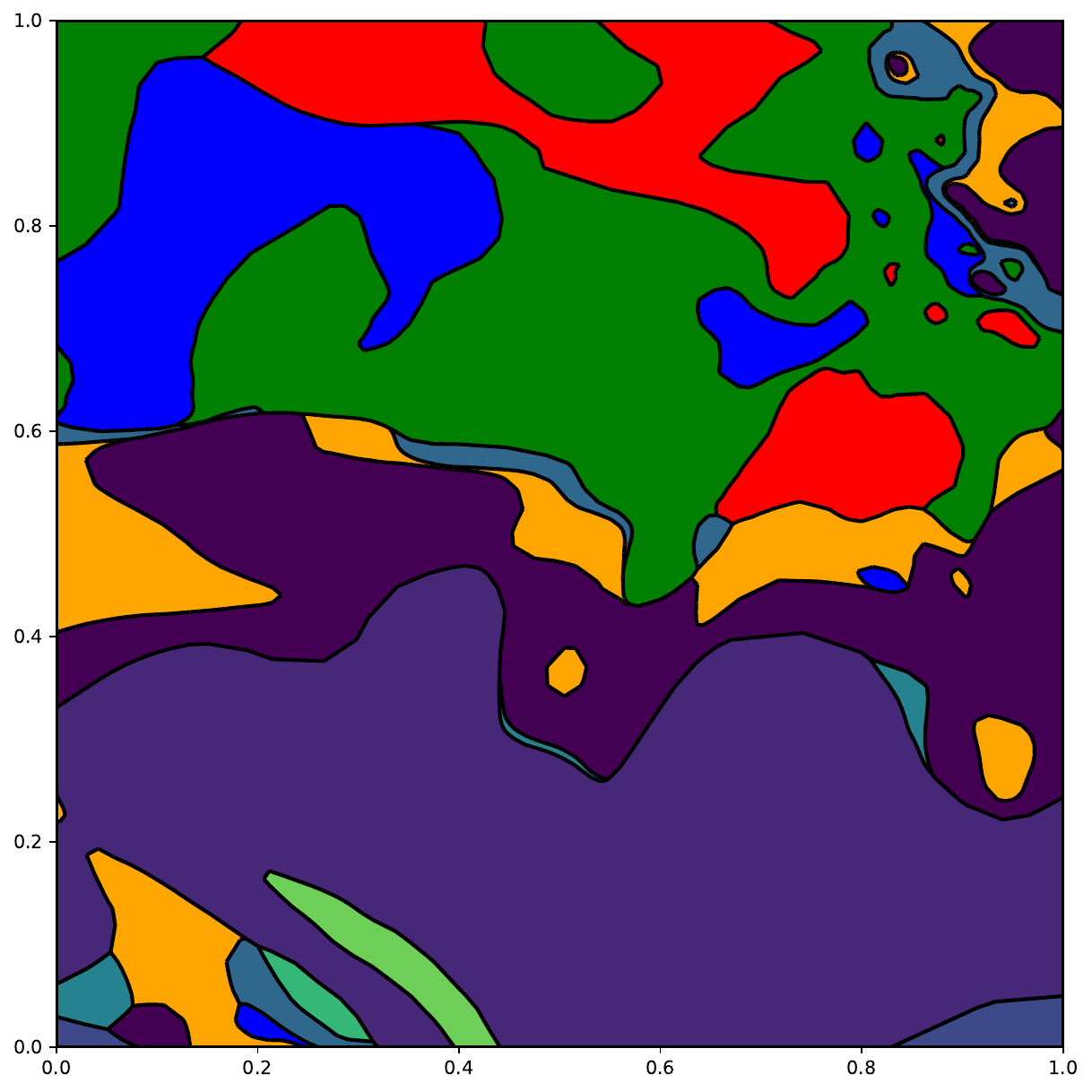}
    \end{minipage}
    \hspace{0.01\textwidth}
    \begin{minipage}[t]{0.31\textwidth}
        \centering
        \includegraphics[width=\linewidth, height=5cm, trim={0cm 0cm 0cm 0cm}, clip]{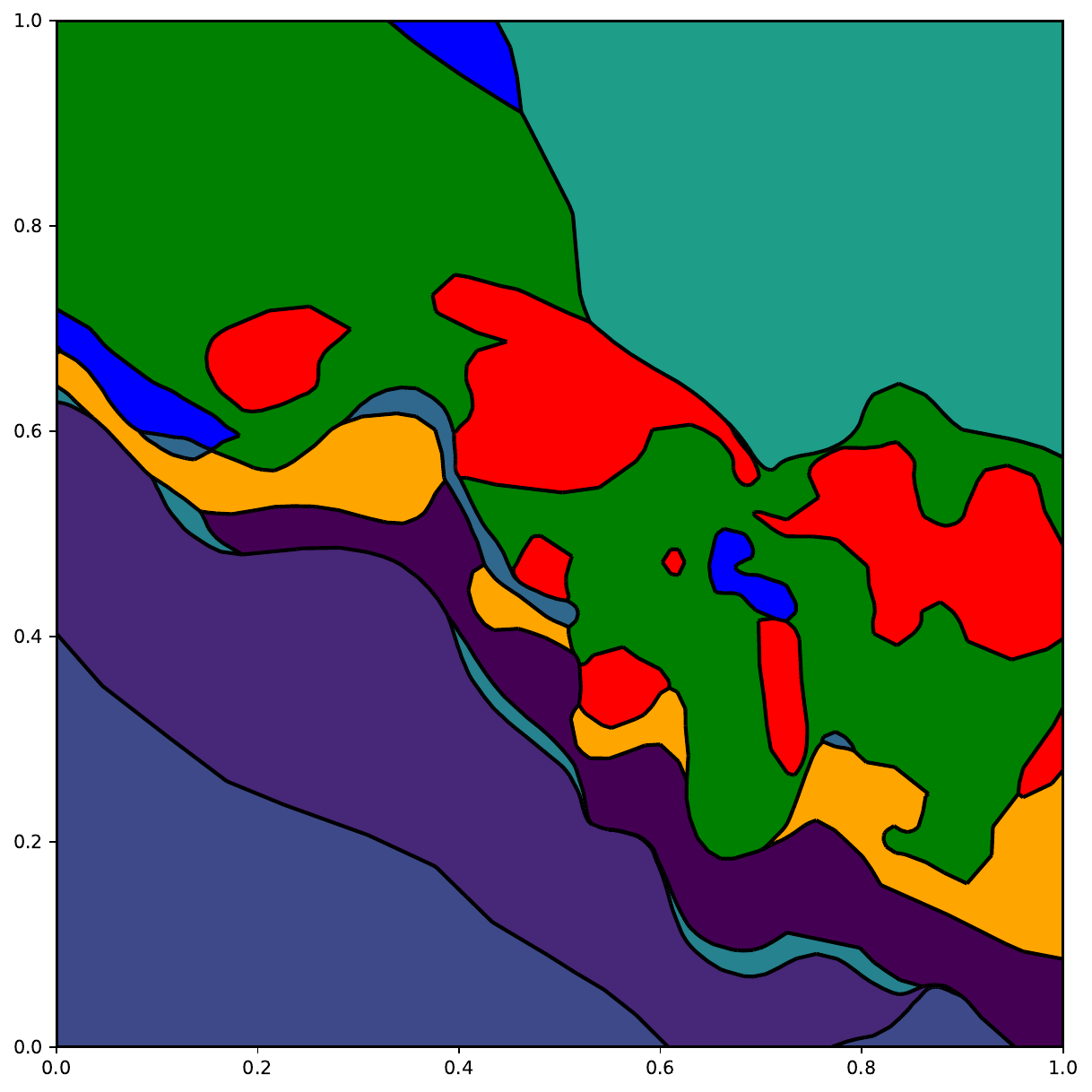}
    \end{minipage}

    \caption{The top row shows slices from a geological model, selected for their complexity, sampled at a resolution 1000x1000 from the neural network. The bottom row shows the 2D polygoniser's result.}
    \label{fig:2Dresults}
\end{figure*}

\begin{figure}[!htpb]
    \centering
    \parbox{0.49\columnwidth}{
        \centering
        \includegraphics[width=\linewidth]{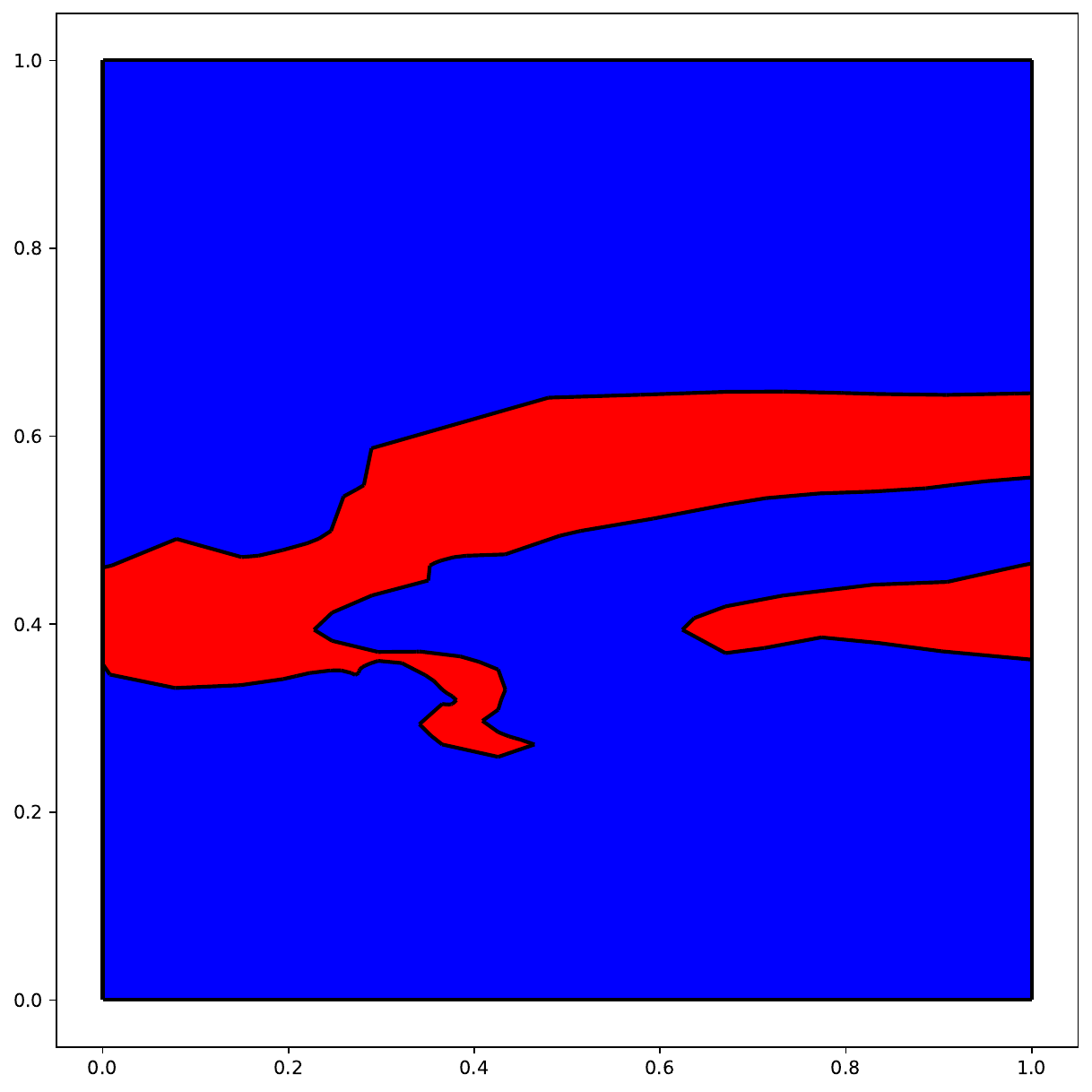}
    }
    \hfill
    \parbox{0.49\columnwidth}{
        \centering
        \includegraphics[width=\linewidth]{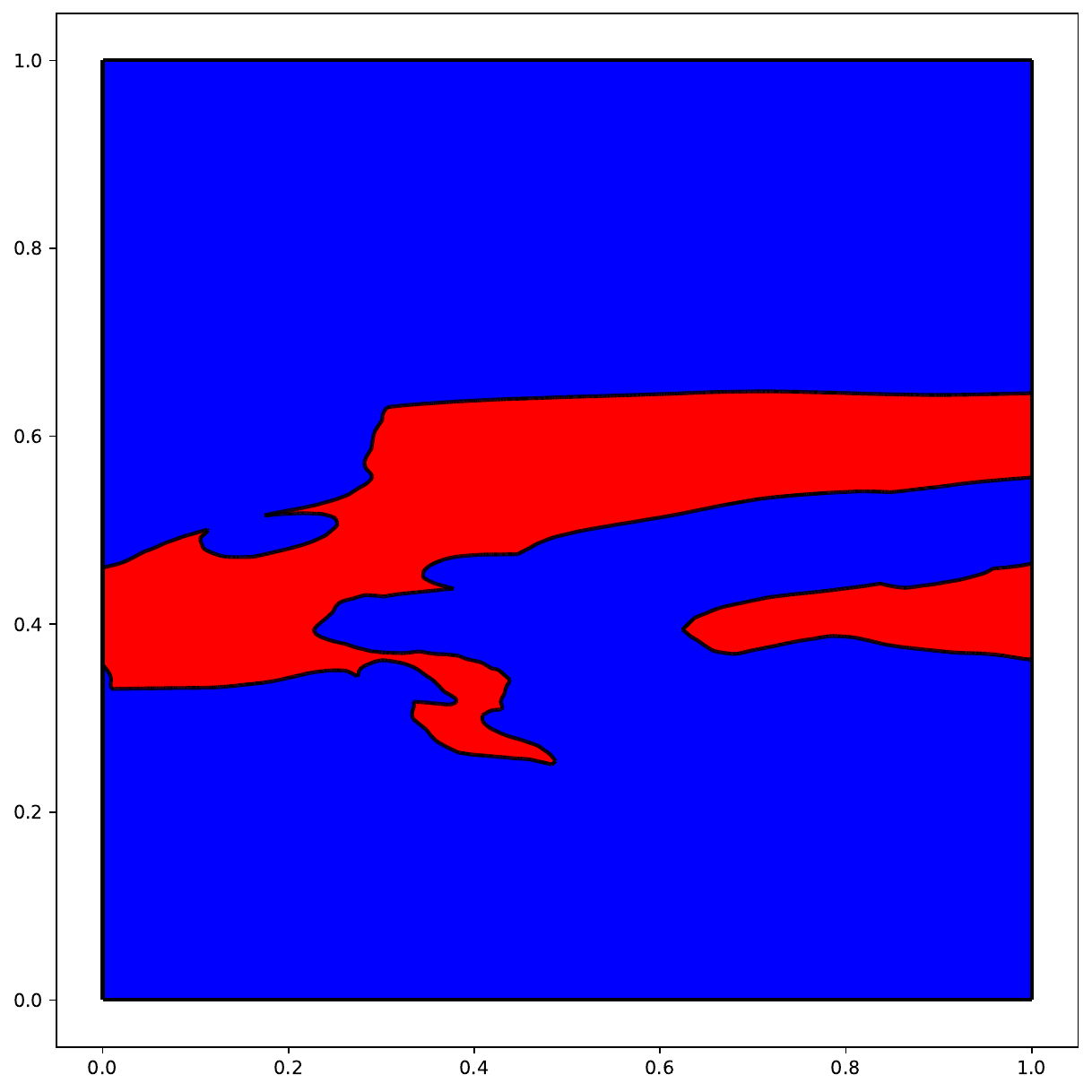}
    }

    \vspace{0.2cm} 

    \parbox{0.49\columnwidth}{
        \centering
        \includegraphics[width=\linewidth]{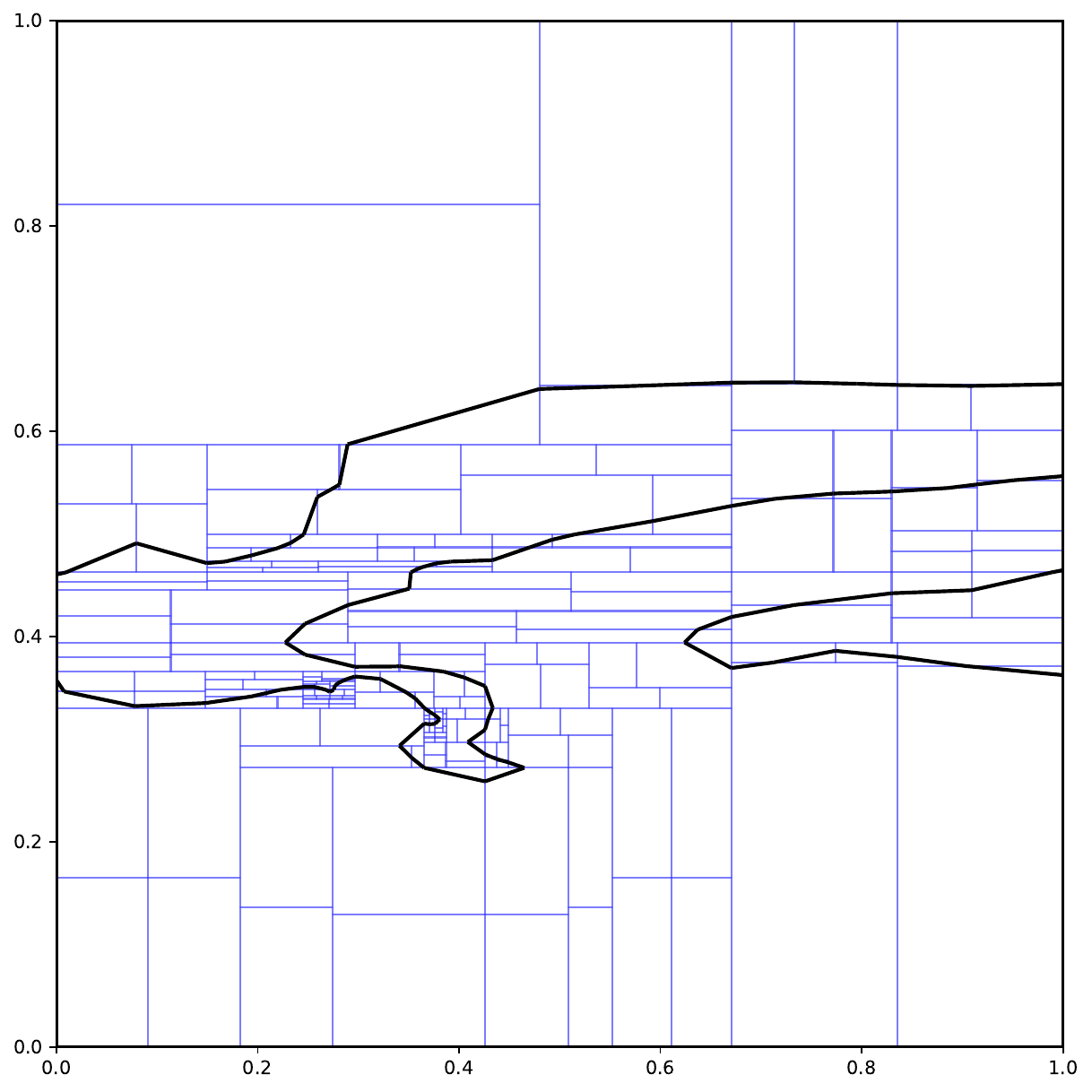}
    }
    \hfill
    \parbox{0.49\columnwidth}{
        \centering
        \includegraphics[width=\linewidth]{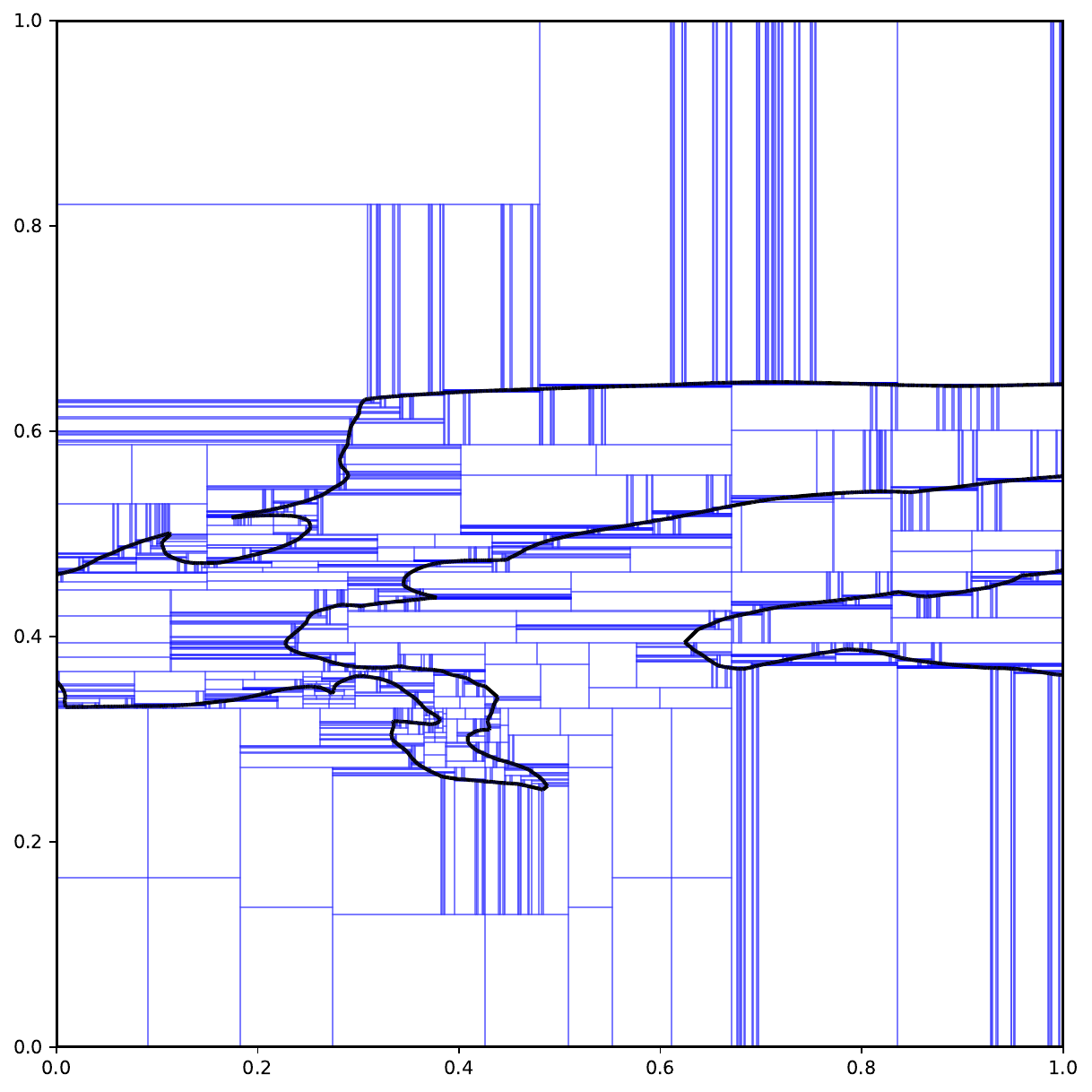}
    }

    \caption{The first column showcases the output of the 2D polygoniser when no geometric criteria were set. The second column shows the output when the geometric criteria threshold was set to 1e-3.}
    \label{fig:gc}
\end{figure}

In this section, we present the effectiveness of the boundary extraction of the 2D polygoniser. We use geological domain data to create an implicit representation\footnote{Due to the proprietary nature of the data, neural network architecture and training details, these can not be released publicly.} $f: \mathbb{R}^3 \to [0,1]^k$ as explained in the previous sections.
The use of geological data shows the real-world effectiveness of our method. Moreover, its complex nature tests heavily for topological consistency and adaptive behaviour, making it a fertile evaluation ground.

\subsection{Adaptive Behaviour}
As noticed in the first column of Figure~\ref{fig:gc}, the adaptive subdivision criteria and verification mechanism allow the algorithm to focus on complex areas and quickly exit topologically uninteresting areas like big expanses of the same domain. Specifically, the ability to subdivide in either direction and choose the point of subdivision based on precise topological information about the borders of a rectangle proves very effective at discerning underlying complexity. The verification level parameter $V_n$ helps capture details where the topology can give deceiving signals (i.e. signalling a \texttt{Polygonisable} class), leading to missing geometry.

\subsection{Topological Consistency}
The 1D root-finder's precision at finding class transitions alongside strict honouring of these 1D topological roots in 2D when subdividing enables features like long smooth veins, no holes, and no self-intersections, as visually observed in Figure~\ref{fig:2Dresults}.
Another important aspect of maintaining topological consistency is reliably finding the intersection point of three domains. This removes any ambiguity associated with the boundaries originating from that triple junction point, as observed in Figure~\ref{fig:2Dresults}.

\subsection{Geometric Criteria}

The algorithm's ability to impose a geometric condition on each edge approximating a class boundary allows a thorough representation of detail. The geometric criteria are only imposed after we are satisfied with the topological integrity of the area. Figure~\ref{fig:gc} displays the change in boundaries after geometric criteria are imposed. We can also observe the additional subdivisions of existing rectangles (also in an adaptive way) for providing more detail to edges that did not meet the stringent geometric criteria threshold $\delta$.

\section{Conclusion and future work}
This work introduces a novel boundary extraction method from multi-class implicit representations in 2D. The algorithm is developed from the ground up to cater for topologically consistent solutions with no holes and self-intersections. Furthermore, geometric constraints can be imposed on the edges, approximating the boundaries for finer detail.

In future work, this 2D algorithm will be used for surface extraction from multi-class implicit representations in 3D, taking an analogous role to the 1D root-finder's role in 2D. This is also the reasoning behind stringent topological consistency requirements from the 2D algorithm.

\clearpage
\newpage
\bibliographystyle{plain}
\bibliography{ref}
\clearpage
\newpage

\appendix

\section{Algorithm: One-Dimensional Root-Finding}

\begin{algorithm}[]
\footnotesize
\caption{One-Dimensional Root-Finding}

\SetKwProg{Fn}{def}{:}{}
\SetKwFunction{Classify}{classify\_interval}
\SetKwFunction{Subdivide}{subdivide\_interval}
\SetKwFunction{Verify}{verify\_intervals}
\SetKwFunction{Gradient}{gradient\_localisation}
\SetKwFunction{Record}{record\_root\_and\_transition}
\SetKwFunction{IntervalLength}{interval\_length}
\SetKwFunction{Enqueue}{enqueue}
\SetKwFunction{Dequeue}{dequeue}
\SetKwIF{If}{ElseIf}{Else}{if}{:}{elif}{else:}{}%
\SetKwFor{While}{while}{:}{}
\SetKwFor{ForEach}{for}{in}{}
\SetAlgoNoEnd

\Input{
    \begin{itemize}
        \item Interval $I_{\text{root}} = [x_{\text{start}}, x_{\text{end}}]$
        \item Projection axis (e.g., $x$ or $y$) and fixed dimension value
        \item Verification depth $V_n$
        \item Interval limit $\epsilon$
    \end{itemize}
}
\Output{Positions of roots and class transitions}

\Fn{root\_finder(interval, axis, fixed\_value, verification\_depth, interval\_limit)}{
    Qv $\leftarrow$ []\;
    Qn $\leftarrow$ []\;
    
    C $\leftarrow$ \Classify{interval, axis, fixed\_value}\;
    \If{C \textbf{in} \{"consistent", "potential root"\}}{
        \Enqueue{Qv, interval}\;
    }
    \Else{
        \Enqueue{Qn, interval}\;
    }
    
    \While{Qv \textbf{or} Qn}{
        \If{Qv}{
            I $\leftarrow$ \Dequeue{Qv}\;
            
            \If{\IntervalLength{I} $<$ interval\_limit}{
                \textbf{continue}\;
            }
            
            S $\leftarrow$ \Subdivide{I, $V_n$}\;
            V $\leftarrow$ \Verify{S, axis, fixed\_value}\;
            
            \If{V}{
                \If{C == "potential root"}{
                    R $\leftarrow$ \Gradient{I, axis, fixed\_value}\;
                    \Record{R}\;
                }
                \textbf{continue}\;
            }
            \Else{
                \Enqueue{Qn, I}\;
            }
        }
        \Else{
            I $\leftarrow$ \Dequeue{Qn}\;
            
            \If{\IntervalLength{I} $<$ interval\_limit}{
                \textbf{continue}\;
            }
            
            left, right $\leftarrow$ \Subdivide{I, 1}\;
            
            \ForEach{child \textbf{in} left, right}{
                C\_child $\leftarrow$ \Classify{child, axis, fixed\_value}\;
                \If{C\_child \textbf{in} \{"consistent", "potential root"\}}{
                    \Enqueue{Qv, child}\;
                }
                \Else{
                    \Enqueue{Qn, child}\;
                }
            }
        }
    }
}
\label{alg:1D_root_finder}
\end{algorithm}

\end{document}